\documentclass[sigconf]{acmart}

\AtBeginDocument{%
  \providecommand\BibTeX{{%
    \normalfont B\kern-0.5em{\scshape i\kern-0.25em b}\kern-0.8em\TeX}}}

\setcopyright{acmcopyright}
\copyrightyear{2020}
\acmYear{2020}
\acmDOI{10.1145/3394486.3403071}

\acmConference[KDD '20] {26th ACM SIGKDD Conference on Knowledge Discovery and Data Mining}{August 23--27, 2020}{Virtual Event, USA}
\acmBooktitle{26th ACM SIGKDD Conference on Knowledge Discovery and Data Mining (KDD '20), August 23--27, 2020, Virtual Event, USA}
\acmPrice{15.00}
\acmISBN{978-1-4503-7998-4/20/08}

\usepackage{epigraph}

\usepackage{fancyhdr}

\DeclareMathOperator*{\argmax}{argmax}

\def\r{\mathbf{r}}
\def\x{\mathbf{x}}

\def\y{\mathbf{y}}
\def\z{\mbox{\boldmath$z$}}

\def\v{\mathbf{v}}

\newtheorem{theorem}{Theorem}

\definecolor{Gray}{gray}{0.85}
\definecolor{LightCyan}{rgb}{0.88,1,1}

\usepackage{color}

\newcommand{\blue}{\color{black}}

\newcommand\blfootnote[1]{%
  \begingroup
  \renewcommand\thefootnote{}\footnote{#1}%
  \addtocounter{footnote}{-1}%
  \endgroup
}

\begin{document}
\fancyhead{}

\title{Adversarial Infidelity Learning for Model Interpretation}



\author{Jian Liang$^{1*}$, Bing Bai$^{1*}$, Yuren Cao$^1$, Kun Bai$^1$, Fei Wang$^2$}
\affiliation{%
  \institution{$^1$Cloud and Smart Industries Group, Tencent, China\\$^2$Department of Population Health Sciences, Weill Cornell Medicine, USA}
}
\email{{joshualiang, icebai, laurenyrcao, kunbai}@tencent.com}
\email{few2001@med.cornell.edu}







\renewcommand{\authors}{Jian Liang, Bing Bai, Yuren Cao, Kun Bai, Fei Wang}

\begin{abstract}
Model interpretation is essential in data mining and knowledge discovery. It can help understand the intrinsic model working mechanism and check if the model has undesired characteristics. A popular way of performing model interpretation is Instance-wise Feature Selection (IFS), which provides an importance score of each feature representing the data samples to explain how the model generates the specific output. In this paper, we propose a Model-agnostic Effective Efficient Direct (MEED) IFS framework for model interpretation, mitigating concerns about {\blue sanity, combinatorial shortcuts, model identifiability, and information transmission}. Also, we focus on the following setting: using selected features to directly predict the output of the given model, which serves as a primary evaluation metric for model-interpretation methods. Apart from the features, we involve the output of the given model as an additional input to learn an explainer based on more accurate information. To learn the explainer, besides fidelity, we propose an Adversarial Infidelity Learning (AIL) mechanism to boost the explanation learning by screening relatively unimportant features. Through theoretical and experimental analysis, we show that our AIL mechanism can help learn the desired conditional distribution between selected features and targets. Moreover, we extend our framework by integrating efficient interpretation methods as proper priors to provide a warm start. Comprehensive empirical evaluation results are provided by quantitative metrics and human evaluation to demonstrate the effectiveness and superiority of our proposed method. Our code is publicly available online at {\blue \url{https://github.com/langlrsw/MEED}.} \blfootnote{$*$ Equal contributions from both authors.}
\end{abstract}




\maketitle

\begin{figure*}[h]
  \centering
  \includegraphics[width=0.8\linewidth,height=5.5cm]{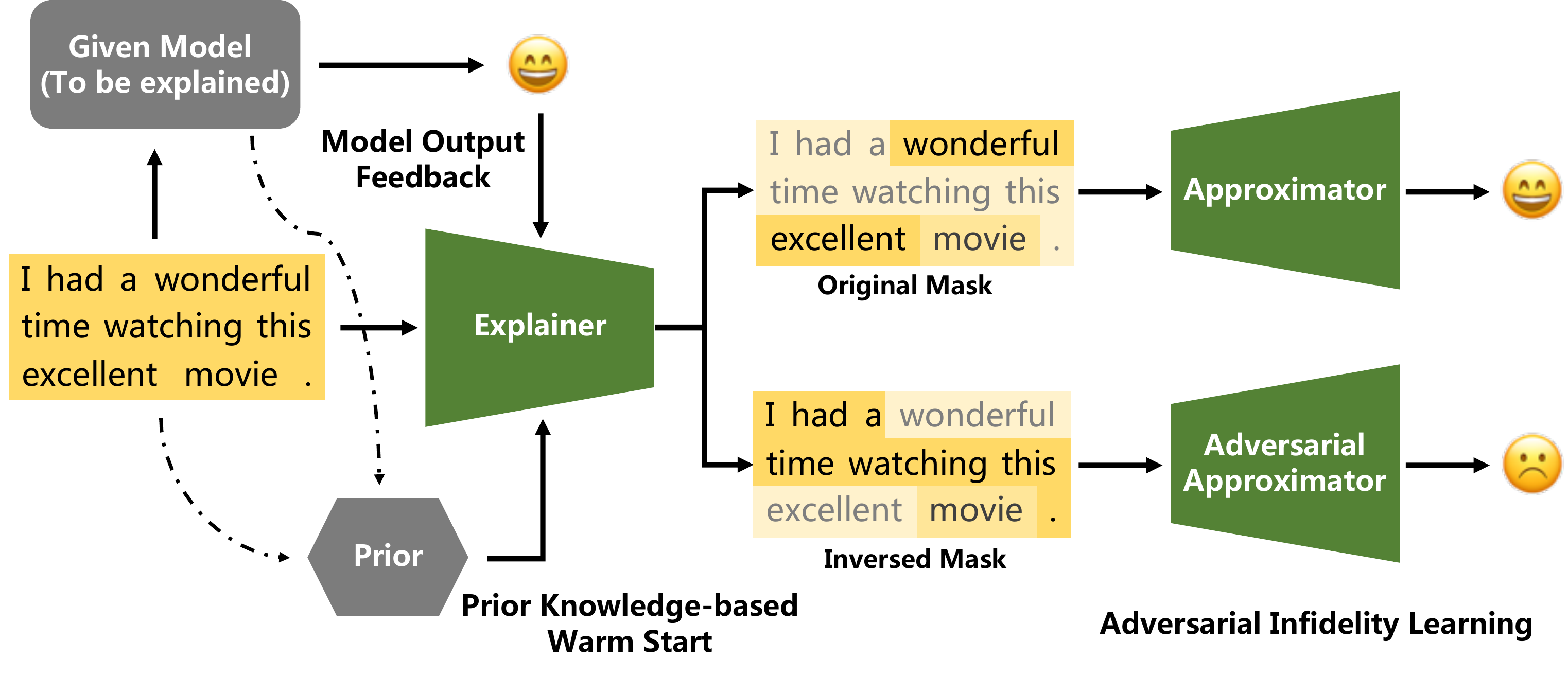}
  \caption{The architecture of our proposed framework. Taking a sentence as an example, we train an explainer to select important words and an approximator to predict the output of the original model. The model output is also used as an input for the explainer. The AIL module trains the explainer to render the adversarial approximator cannot predict the model output well based on the unselected words. As an extension, efficient interpretation methods, \emph{e.g.,} gradient-based methods, can be integrated to provide a warm start. Best view in color.
  }
  \label{fig:framwork}
\end{figure*}

\section{Introduction}
 

The interpretation of data-driven models explains their input-output relationship, which provides information about whether the models admit some undesired characteristics, and thus can guide people to use, debug, and improve machine learning models. The model interpretation has an increasing demand in many real-applications, including medicine~\cite{wang2019should}, security~\cite{chakraborti2019explicability}, and criminal justice~\cite{lipton2016mythos}.  

Existing research on model interpretation can be categorized into \emph{model-specific methods} and \emph{model-agnostic methods}. Model-specific methods take advantage of the knowledge of the model itself to assist explanations, such as gradient-based methods for neural networks, whereas model-agnostic methods can explain any black-box system. Instance-wise Feature Selection~(IFS) is a well known model-agnostic interpretation method. It produces an importance score of each feature for representing a data sample~\cite{du2018techniques}, which indicates how much each feature dominates the model's output.
For this kind of approach, desired properties for ideal explanations (feature importance scores) are as follows.
\begin{itemize}
    \item Expressiveness: the number of features with relatively high scores should be small~\cite{ribeiro2016should}.
    \item Fidelity: the model output should primarily depend on high-score features~\cite{ribeiro2016should,chen2018learning,bang2019explaining,hara2019feature,hooker2019benchmark,yeh2019fidelity,schwab2019cxplain,khakzar2019explaining}.
    \item Low sensitivity: feature scores should be robust against adversarial attacks~\cite{yeh2019fidelity,NIPS2019_9511,NIPS2019_8558,zhang2018interpretable}.
    \item Sanity: feature scores should be dependent of the model~\cite{adebayo2018sanity}.
\end{itemize}

Recent research for IFS-based model explanation can be divided into \emph{(local/global) feature attribution methods}~\cite{ancona2017unified,yeh2019fidelity}\footnote{In this paper, the definitions of global and local explanations follow the description of Ancona~\emph{et al.}~\cite{ancona2017unified} and Yeh~\emph{et al.}~\cite{yeh2019fidelity}, and distinct from that of Plumb~\emph{et al.}~\cite{plumb2018model}.} and \emph{direct model-interpretation~(DMI) methods}. Local feature attribution methods provide some sensitivity scores of the model output concerning the changes of the features in the neighborhood. In contrast, global feature attribution methods directly produce the amount of change of the model output given changes of the features. Other than providing the change of the model output, DMI is a more straightforward approach to select features and use a model to approximate the output of the original black-box model~\cite{chen2018learning,sundararajan2017axiomatic}.

In this paper, we attempt to tackle the DMI problem. \emph{When given a data sample and the model to be explained, what features does the model use primarily to generate the output?} A straightforward approach is to develop a feature attribution network~(which we refer to as the explainer) to produce a soft/hard mask to highlight essential features, and a prediction network~(which we refer to as the approximator) to approximate the output of the original model~\cite{chen2018learning}. {\blue 
However, this straightforward approach may cause the effectiveness and efficiency related concerns in the following.

\begin{itemize}
    \item \textbf{Sanity problem}~\cite{adebayo2018sanity}: a mask may be irrelevant to the original model, but only relate to the features of a specific sample. As a consequence, the selected features of the trained explainer may be different with those truly used by the original model, which is not expected in interpreting the model.
    \item \textbf{Combinatorial shortcuts problem}: the entries of the mask may not select good features, but rather act as additional features themselves for better approximation performances~\cite{jain2019attention,wiegreffe2019attention}, because it is a function of all the input features. For example, the explainer could choose to mask out the first half of the features for positive samples, and the second half of the features for negative samples. The approximator can utilize this pattern to predict the target while completely ignore whether good features are selected.
    \item \textbf{Model identifiability problem}: similar approximation performances can be achieved by different groups of feature. It is difficult to decide which group is the best.
    \item \textbf{Information transmission problem}~\cite{park2019fast}: it is difficult to transmit effective supervised information to the explainer, because the mask is unsupervised.
\end{itemize}

}


To address these issues, we propose a Model-agnostic Effective Efficient Direct (MEED) model-interpretation method for the DMI problem. The overall architecture of our proposed framework is presented in Figure~\ref{fig:framwork}. The major components include model output feedback, adversarial infidelity learning, and prior knowledge-based warm start, which we describe as follows. 


Firstly, we propose to enrich the input information of the explainer to boost the effectiveness and the efficiency of the feature selection process. Existing research treats raw features {\blue only} as the input to the neural network-based explainer~\cite{schwab2019cxplain,chen2018learning,bang2019explaining}. {\blue The absence of the original model to include for the explainer's input may render the mask out of the explainer uncorrelated with the original model
and then cause the sanity problem. Nonetheless, it is not trivial to input a whole model into the neural network-based explainer. Therefore, we propose to incorporate the model output as another input signal. Apart from the sanity problem, the model output can provide rich information for the explainers to select essential features, and make the learning process more precise, especially in applications like regression or representation learning.}
In other words, the information transmission problem can also be mitigated.

Secondly, we propose to exploit the unselected features for mitigating the combinatorial shortcuts and model identifiability problems. Inspired by Hooker~\emph{et al.}~\cite{hooker2019benchmark}, we attempt to achieve an auxiliary goal that \emph{the unselected features should contain the least useful information}.
To achieve this, we propose an Adversarial Infidelity Learning (AIL) mechanism. Specifically, we develop another approximator that learns to approximate the original model output using the unselected features. Then our explainer learns to select features to minimize such approximation accuracy. The learning processes run alternately. {\blue Intuitively, the convergence of such an adversarial learning process will render the masks uncorrelated with the model output, and then can mitigate the combinatorial shortcuts problem. On the other hand, this learning process exploits the unselected features, which are often (at least relatively) ignored, to introduce additional supervised information for a certain group of selected features, and then can improve model identifiability. These properties are demonstrated by our theoretical analysis and experimental results.}

Finally, we extend our framework to further mitigate the information transmission problem by integrating prior knowledge. Specifically, we integrate explanations provided by efficient interpretation methods as priors to provide a warm start. The constraints of the priors fade out when the number of training epochs grows to learn a more powerful explainer by the end-to-end framework.

{\blue We follow Chen~\cite{chen2018learning} to perform a predictive evaluation to see whether the selected features contribute to sufficient approximate accuracy.} Comprehensive empirical evaluation results on four real-world benchmark datasets are provided with quantitative evaluation metrics and human-evaluations to demonstrate the effectiveness and superiority of our proposed method. Moreover, we validate our method on a real-world application: teenager/adult classification based on mobile sensor data from $5$ million of Tencent users who play the popular \emph{Honor of Kings, a.k.a. Arena of Valor } game. 


    
     


\section{Related Works}
\label{sec:related}

Model interpretation methods based on IFS can be categorized into local/global methods as introduced in the introduction. 
Local methods includes 1) gradient-based methods, such as Gradient (Grad)~\cite{simonyan2013deep} and Guided Back-Propagation~\cite{springenberg2014striving}, 2) sampling-based methods, \emph{i.e.,} perform sensitivity analysis by sampling points around the given data sample, such as LIME~\cite{ribeiro2016should}, kernel SHAP~\cite{lundberg2017unified} and CXPlain~\cite{schwab2019cxplain}, and 3) hybrid methods, such as SmoothGrad~\cite{smilkov2017smoothgrad}, Squared SmoothGrad~\cite{smilkov2017smoothgrad}, VarGrad~\cite{adebayo2018local}, and INFD~\cite{yeh2019fidelity}. On the other hand, global methods include Gradient $\times$ Input~\cite{shrikumar2017learning}, Integrated Gradients~\cite{sundararajan2017axiomatic}, DeepLIFT~\cite{shrikumar2017learning} and LRP~\cite{bach2015pixel}, among others. 
These methods do not directly tackle the DMI problem.

For the DMI problem, being inherently interpretable, tree-~\cite{schwab2015capturing} and rule-based~\cite{andrews1995survey} models have been proposed to approximate the output of a complex black-box model with all features. The models themselves provide explanations, including feature importance. However, they may lack the ability for accurate approximations when the original given model is complex. Recently, L2X~\cite{chen2018learning} and VIBI~\cite{bang2019explaining} have been proposed as 
variational methods to learn a neural network-based 
approximator based on the selected features. The unselected features are masked out by imputing zeros. VIBI improves L2X to encourage the briefness of the learned explanation by adding a constraint for the feature scores to a global prior (in contrast, our priors are conditioned on each sample). Since L2X and VIBI only input features to their explainers, and they directly select features to approximate the model out, therefore, they both may suffer from the {\blue sanity, combinatorial shortcuts, model identifiability, model identifiability, and information transmission} problems.

 


In contrast, {\blue our method tackle these problems for DMI}
by leveraging more comprehensive information from the model output, the proposed adversarial infidelity learning mechanism, and the proposed prior-knowledge integration.

Apart from model interpretation, we note that Zhu~\emph{et al.}~\cite{zhu2019adversarial} proposed an adversarial attention network. However, their objective is to eliminate the difference in extracted features for different learning tasks, which is different from ours. 
We also note that Yu~\emph{et al.}~\cite{yu2019rethinking} proposed a similar approach with ours working on selective rationalization, whereas ours is targeted at model interpretation. Specifically, they proposed an introspection mask generator to treat the prediction of their label predictor as another input signal, in order to disentangle the information about the correct label. Whereas our model-output-feedback module is designed to mitigate the sanity and the information transmission problems of model interpretation. Additionally, they also proposed an adversarial mechanism to provide explicit control over unselected features, aiming to avoid the failure of the cooperative game between their mask generator and label predictor. Whereas this paper points out that such failure may be caused by combinatorial shortcuts and proposes a theoretical guaranteed method to mitigate the combinatorial shortcuts problem. Finally, this paper further proposes a prior knowledge-based warm start module to mitigate the information transmission problem. Moreover, Chang~\emph{et al.}~\cite{chang2019game} proposed a class-wise adversarial rationalization method to select features supporting counterfactual labels. Chang~\emph{et al.}~\cite{chang2020invariant} extended the work of Yu~\emph{et al.}~\cite{yu2019rethinking} to select features which are invariant for different environments/domains, assuming the environment/domain labels are given.



\section{Methodology}
\label{sec:method}

In this section, we present the detailed methodology of our method. First, we define the notations and problem settings of our study.

Consider a dataset $\mathcal{D} = \{\x^i\}_{i=1}^n$ consisting of $n$ independent samples. For the $i$th sample, $\x^i\in\mathcal{X}\subset\mathbb{R}^d$ is a feature vector with $d$ dimensions, $\y^i=M(\x^i)\in\mathcal{Y}\subset\mathbb{R}^c$ is the output vector of a given data-driven model $M\in\mathcal{M}$~(note that $y^i$ may be different from the true label of the sample). The conditional output distribution $p(\y\mid \x)$ is determined by the given model. For classification tasks, $c$ is the number of classes. 

We do not assume the true label of each feature vector is available for training or inference. We develop a neural network-based IFS explainer $E$, which outputs a feature-importance-score vector $\z\in \mathcal{Z}\subset \mathbb{R}^d$ for a data sample $\x$ and the model $M$. As discussed by Yeh~\emph{et al.}~\cite{yeh2019fidelity}, the explainer should be a mapping that $E:\mathcal{X}\times \mathcal{M}\rightarrow\mathcal{Z}$. Since it is not trivial to treat an arbitrary model in $\mathcal{M}$ as an input to a neural network, we compromise by involving the model output as an alternative such that $E:\mathcal{X}\times \mathcal{Y}\rightarrow\mathcal{Z}$. 
We select top-$k$ features according to $\z$, where $k$ is a {\blue user-defined} parameter. The indices of $k$ selected features are denoted by $S\subset \{1,\ldots,d\}$. For a feature vector $\x$, the selected features are denoted by $\x_S$, whereas the unselected features are denoted by $\x_{\bar{S}}$. Throughout the paper, we denote $[k']$ as the index set $\{1,2,\ldots,k'\}$ for some integer $k'$.

The goal of our learning approach is to train a neural network-based explainer $E$ over the dataset $\mathcal{D}$ and then generalize it to a testing set to {\blue see whether the selected features contribute to sufficient approximate accuracy.} The quantitative evaluations of the explainer are described in Section~\ref{subsec:PE}. 

\subsection{Our Framework}\label{subsec:framework}

The architecture of our framework is illustrated in Fig.~\ref{fig:framwork}. We explain a given model by providing IFS for each specific data sample. The IFS is embodied as a feature attribution mask provided by a learned explainer with the features and the model output of the data sample as inputs. We train an approximator to use selected/masked features to approximate the model output. We also train an adversarial approximator to use unselected features to approximate the model output, and then train the explainer to select features to undermine the approximation, which is referred to as the AIL mechanism. As an extention, integrating efficient model-interpretation methods is also introduced to provide a warm start.


\subsection{Adversarial Infidelity Learning}\label{subsec:AIL}

As discussed in the introduction, a straightforward approach to optimize the selection indices $S$ is directly maximizing the mutual information $I(\x_S;\y)$ between selected features $\x_S$, and the model output $\y$~\cite{chen2018learning,bang2019explaining}. {\blue To tackle the combinatorial shortcuts and model identifiability problems}, we propose an auxiliary objective: minimizing the mutual information $I(\x_{\bar{S}};\y)$ between unselected features $\x_{\bar{S}}$ and the model output $\y$. Because compared with the selected features, the unselected features should contain {\blue less} useful information. Therefore, the basic optimization problem for $S$ is:
\begin{equation}\label{eq:basic_AIL}
    \max_S \ I(\x_S;\y) - I(\x_{\bar{S}};\y)
    \ \ \mbox{s.t.} \ S\sim E(\x,\y).
\end{equation}
We can be guided by the Theorem~\ref{th:mi} to optimize the explainer.
\begin{theorem}\label{th:mi}
Define
\begin{equation}\label{eq:th_opt_mi}
    S^* = \argmax_S \mathbb{E}[\log p(\y\mid \x_S) - \log p(\y\mid \x_{\bar{S}})],
\end{equation}
where the expectation is over $p(\y\mid\x)$.
Then $S^*$ is a global optimum of Problem~\eqref{eq:basic_AIL}. Conversely, any global optimum of Problem~\eqref{eq:basic_AIL} degenerates to $S^*$ almost surely over $p(\x,\y)$. 
\end{theorem}
The proof is deferred to Appendix~\ref{subsec:proof_1}. {\blue Problem~\eqref{eq:basic_AIL} and Theorem~\ref{th:mi} show that the auxiliary objective exploits $\x_{\bar{S}}$ to involve additional supervised information, and then improves model identifiability.}

According to Theorem~\ref{th:mi}, we deveop an approximator $A_s:\mathcal{X}\rightarrow \mathcal{Y}$ to learn the conditional distribution $p(\y \mid \x_S)$. We achieve this by optimizing a variational mapping: $\x_S\rightarrow q_s(\y\mid\x_S)$ to let $q_s(\y\mid\x_S)$ approximate $p(\y \mid \x_S)$. We define $q_s(\y\mid\x_S)\propto \exp(-\ell_s(\y,A_s(\tilde{\x}_S)))$, where $\ell_s$ denotes the loss function corresponding to the conditional distribution $p(\y \mid \x_S)$ (\emph{e.g.,} mean square error for Gaussian distribution, and categorical cross entropy for categorical distribution), and $\tilde{\x}_S\in\mathcal{X}$ which is defined as: $(\tilde{\x}_S)_j=x_j$ if $j\in S$ and $(\tilde{\x}_S)_j=0$ otherwise. We let $q_m(\y\mid \tilde{\x}_S)$ denote the output distribution of $M(\tilde{\x}_S)$. {\blue We approximate $\y$ by $A_s$ instead of $M$, because} as discussed by Hooker~\emph{et al.}~\cite{hooker2019benchmark}, $p(\y \mid \x_S)\neq q_m(\y\mid \tilde{\x}_S)$, then $A_s(\tilde{\x}_S)$ {\blue may approximate more accurate than $M(\tilde{\x}_S)$ does}. 

Similarly, we develop another approximator $A_u:\mathcal{X}\rightarrow \mathcal{Y}$ to learn $q_u(\y\mid \x_{\bar{S}})
\propto \exp(-\ell_u(\y,A_u(\tilde{\x}_{\bar{S}})))$ to approximate $p(\y\mid\x_{\bar{S}})$. Then we can show that Problem~\eqref{eq:basic_AIL} can be relaxed by maximizing variational lower bounds and alternately optimizing: 
\begin{equation}\label{eq:basic_AIL_relax_pos}
    \max_{A_s,A_u} \ \mathbb{E}[\log q_s(\y\mid \x_S) + \log q_u(\y\mid \x_{\bar{S}}) ]
    \ \ \mbox{s.t.} \ S\sim E(\x,\y),
\end{equation}
\begin{equation}\label{eq:basic_AIL_relax_neg}
    \max_{E} \ \mathbb{E}[\log q_s(\y\mid \x_S) - \log q_u(\y\mid \x_{\bar{S}}) ]
    \ \ \mbox{s.t.} \ S\sim E(\x,\y).
\end{equation}
{\blue First, Problem~\eqref{eq:basic_AIL_relax_pos} is  optimized} to learn $q_s(\y\mid \x_S)$ and $q_u(\y\mid \x_{\bar{S}})$ to approximate $p(\y\mid \x_S)$ and $p(\y\mid \x_{\bar{S}})$, respectively. Then Problem~\eqref{eq:basic_AIL_relax_neg} is optimized to learn the explainer $E$ to find good explanations according to Theorem~\ref{th:mi}. 
{\blue Since 1) $q_u(\y\mid \x_{\bar{S}})$ is maximized by optimizing $A_u$ and then minimized by optimizing $E$, which is an \emph{adversarial learning} process, and 2) minimizing $q_u(\y\mid \x_{\bar{S}})$ represents \emph{infidelity}, \emph{i.e.,} undermining performance to approximate $\y$ (by excluding selected features), the alternate optimization process can be regarded as an adversarial infidelity learning mechanism.}

Since optimizing Problems~\eqref{eq:basic_AIL_relax_pos} and~\eqref{eq:basic_AIL_relax_neg} for all possible $S$ requires $\binom{d}{k}$ times of computation for the objectives, we follow L2X~\cite{chen2018learning} to apply the Gumbel-softmax trick to approximately sample a $k$-hot vector. Specifically, let $\z=E(\x,\y)$ for a pair of inputs $(\x,\y)$, where for $j\in[d]$, $z_j \geq 0$ and $\sum_jz_j=1$. Then we define the sampled vector $\v\in\mathcal{V}\subset\mathbb{R}^d$, where for a predefined $\tau>0$,
\begin{equation}\label{eq:gumble}
\begin{split}
v_j & = \max_{l\in[k]} \frac{\exp((\log z_j + \xi_j^l)/\tau)}{\sum_{j'=1}^d\exp((\log z_{j'} + \xi_{j'}^l)/\tau)}, \ j\in[d],\\
\xi_j^l &= -\log(-\log u_j^l), \ u_j^l \sim \mbox{Uniform(0,1)}, \ j\in[d],l\in[k].
\end{split}
\end{equation}

Denote the above random mapping by $G:\mathcal{Z}\rightarrow\mathcal{V}$, approximate $\tilde{\x}_S\approx\x \odot G(E(\x,\y))$ and $\tilde{\x}_{\bar{S}}\approx\x \odot (\mathbf{1}^d-G(E(\x,\y)))$, where $\mathbf{1}^d\in\mathbb{R}^d$ with all elements being $1$, and $\odot$ denotes element-wise product. Define the losses for {\blue selected and unselected features}, respectively,
\begin{equation}\label{eq:loss_AIL}
\begin{split}
\mathcal{L}_s &= \frac{1}{n}\sum_{i=1}^n\ell_s(\y^i,A_s(\x^i \odot G(E(\x^i,\y^i))))\\
\mathcal{L}_u &= \frac{1}{n}\sum_{i=1}^n\ell_u(\y^i,A_u(\x^i \odot (\mathbf{1}^d-G(E(\x^i,\y^i)))).
\end{split}
\end{equation}

Then we can relax Problems~\eqref{eq:basic_AIL_relax_pos} and~\eqref{eq:basic_AIL_relax_neg} as
\begin{equation}\label{eq:basic_AIL_relax_pos_final}
    \min_{A_s,A_u} \ \mathcal{L}_s + \mathcal{L}_u,
\end{equation}
\begin{equation}\label{eq:basic_AIL_relax_neg_final}
    \min_{E} \  \mathcal{L}_s - \mathcal{L}_u.
\end{equation}

For inference, one can select the top-$k$ features of $E(\x^t,\y^t)$ for a testing sample $\x^t$, where $\y^t=M(\x^t)$.




\subsection{Theoretical Analysis}

In this section, we first derive variational lower bounds to show the connection between the goal in Eq.~\eqref{eq:basic_AIL} and the realization in Eq.~\eqref{eq:basic_AIL_relax_pos} and~\eqref{eq:basic_AIL_relax_neg}. The derivation also shows that the approximator is possible to be superior to the given model to predict the model output using masked features. We also show that our AIL mechanism can mitigate the combinatorial shortcuts problem.

The variational lower bounds are as follows and derived in Appendix~\ref{subsec:VLB}. 
First, for selected features, we have for any $q_s(\y\mid \x_S)$:
\begin{equation}\label{eq:basic_AIL_relax_proof_mi}
\begin{split}
    &I(\x_S;\y) 
    =\mathbb{E}_{\x}\mathbb{E}_{\y\mid\x}\mathbb{E}_{S\mid\x,\y}\mathbb{E}_{\y\mid\x_S}[\log p(\y\mid\x_S)] + \mbox{Const,}\\
&\mathbb{E}_{\y\mid\x_S}[\log p(\y\mid\x_S)]
\geq \mathbb{E}_{\y\mid\x_S}[\log q_s(\y\mid \x_S)],
\end{split}
\end{equation}
where the equality holds if and only if 
$q_s(\y\mid \x_S)=p(\y\mid\x_S)$. Therefore, if $M(\tilde{\x}_S)$'s output distribution $q_m(\y\mid \tilde{\x}_S)\neq p(\y\mid\x_S)$, it is possible that $\mathbb{E}_{\y\mid\x_S}[\log q_s(\y\mid \x_S)]>\mathbb{E}_{\y\mid\x_S}[\log q_m(\y\mid \tilde{\x}_S)]$, \emph{i.e.,} $A_s(\tilde{\x}_S)$ can be more accurate than $M(\tilde{\x}_S)$ to estimate $\y$.

Similarly, for unselected features, we have for any $q_u(\y\mid \x_{\bar{S}})$:
\begin{equation}\label{eq:basic_AIL_relax_proof_mi_3}
\begin{split}
&I(\x_{\bar{S}};\y) 
    =\mathbb{E}_{\x}\mathbb{E}_{\y\mid\x}\mathbb{E}_{S\mid\x,\y}\mathbb{E}_{\y\mid\x_{\bar{S}}}[\log p(\y\mid\x_{\bar{S}})] + \mbox{Const.}\\
&\mathbb{E}_{\y\mid\x_{\bar{S}}}[\log p(\y\mid\x_{\bar{S}})]
\geq \mathbb{E}_{\y\mid\x_{\bar{S}}}[\log q_u(\y\mid \x_{\bar{S}})]. 
\end{split}
\end{equation}

On the other hand, since $A_s$ actually receives the selected features $\x_{S}$ and the feature-attribution mask $\v$ as inputs, where $\v = G(E(\x,\y))$, what $A_s$ actually learns is the conditional distribution $p(\y\mid \x_{S},\v)$. Through the straightforward learning mentioned in the introduction, \emph{i.e.,} removing $\mathcal{L}_u$ in Eq.~\eqref{eq:basic_AIL_relax_pos_final} and~\eqref{eq:basic_AIL_relax_neg_final},
it could cause the combinatorial shortcuts problem for $A_s$ to learn $p(\y\mid \v)$ only, resulting in the feature selection process $\x_{S}$ meaningless. 
Fortunately, Theorem~\ref{th:ail} shows that our AIL can help to avoid this problem by encouraging the independence between $\v$ and $\y$, then it will be hard for $A_s$ to approximate $\y$ solely by $\v$. Thus, $A_s$ will have to select useful features from $\x$. The proof can be found in Appendix~\ref{subsec:proof_2}.
\begin{theorem}\label{th:ail}
For the optimized problem in Eq.~\eqref{eq:basic_AIL_relax_pos_final} and~\eqref{eq:basic_AIL_relax_neg_final},  the independence between $\v$ and $\y$ is encouraged.
\end{theorem}





\subsection{Extension Considering Prior Knowledge}\label{subsec:PEI}


As described in Section~\ref{subsec:AIL}, the feature attribution layer (the output layer of the explainer) is in the middle of networks. Since it is optimized by an end-to-end learning process, the information transmission is inefficient. Therefore, we propose to involve efficient interpretations as good priors of feature attribution.

Let $\r\in\mathbb{R}^d$ be a feature-importance-score vector generated by another interpretation method $H$ for a sample $\x$ and the model $M$. Assume for $j\in[d]$, $r_j\geq0$ and $\sum_jr_j=1$, which can be easily achieved through a softmax operation. 

Given $\z=E(\x,M(\x))$ for a sample $\x$, we can regard $z_j$ as $p(\delta = j\mid \x,M,E)$ for $j\in[d]$, where $\delta \in[d]$ denotes whether the $j$th feature should be selected. Similarly, we can regard $r_j$ as $p(\delta = j\mid \x,M,H)$. Then assuming conditional independence between the interpretation models $E$ and $H$ given $\delta,\x$ and $M$, we can obtain 
\begin{equation}\label{eq:naive_bayes}
p(\delta = j\mid \x,M,E,H) = \frac{p(\delta = j\mid \x,M,E)p(\delta = j\mid \x,M,H)}{\sum_{j'=1}^dp(\delta = j'\mid \x,M,E)p(\delta = j'\mid \x,M,H)}.
\end{equation}
The derivation details can be found in Appendix~\ref{subsec:proof_3}.

Nonetheless, as we expect that the end-to-end learning process can generate better explanations, the prior explanation should decrease its influence when the number of epochs $m\in\mathbb{Z}_+$ increases. Therefore, we define
\begin{equation}\label{eq:decrease}
\begin{split}
\tilde{\z} := \frac{[p(\delta = j\mid \x,M,E)^mp(\delta = j\mid \x,M,H)]^{1/(m+1)}}{\sum_{j'=1}^d[p(\delta = j'\mid \x,M,E)^mp(\delta = j'\mid \x,M,H)]^{1/(m+1)}}.
\end{split}
\end{equation}
For Eq.~\eqref{eq:gumble}, we replace $\z$ with the re-estimated $\tilde{\z}$ defined in Eq.~\eqref{eq:decrease}.

In addition, we add a constraint for the explanations, learning $\z = E(\x,M(\x))$ to be close to $\tilde{\z}$ for a loss $\ell_e(\cdot,\cdot)$ :
\begin{equation}\label{eq:constraint}
\mathcal{L}_e = \frac{1}{n}\sum_{i=1}^n
\frac{\ell_e(\tilde{\z}^i,  \z^i)}{m+1}.
\end{equation}
The constraint will fade out when the number of epochs $m$ becomes large, and thus only contributes for a warm start. 

\section{Experiments}\label{sec:exp}

\begin{figure*}[h]
  \centering
  \includegraphics[width=0.8\linewidth]{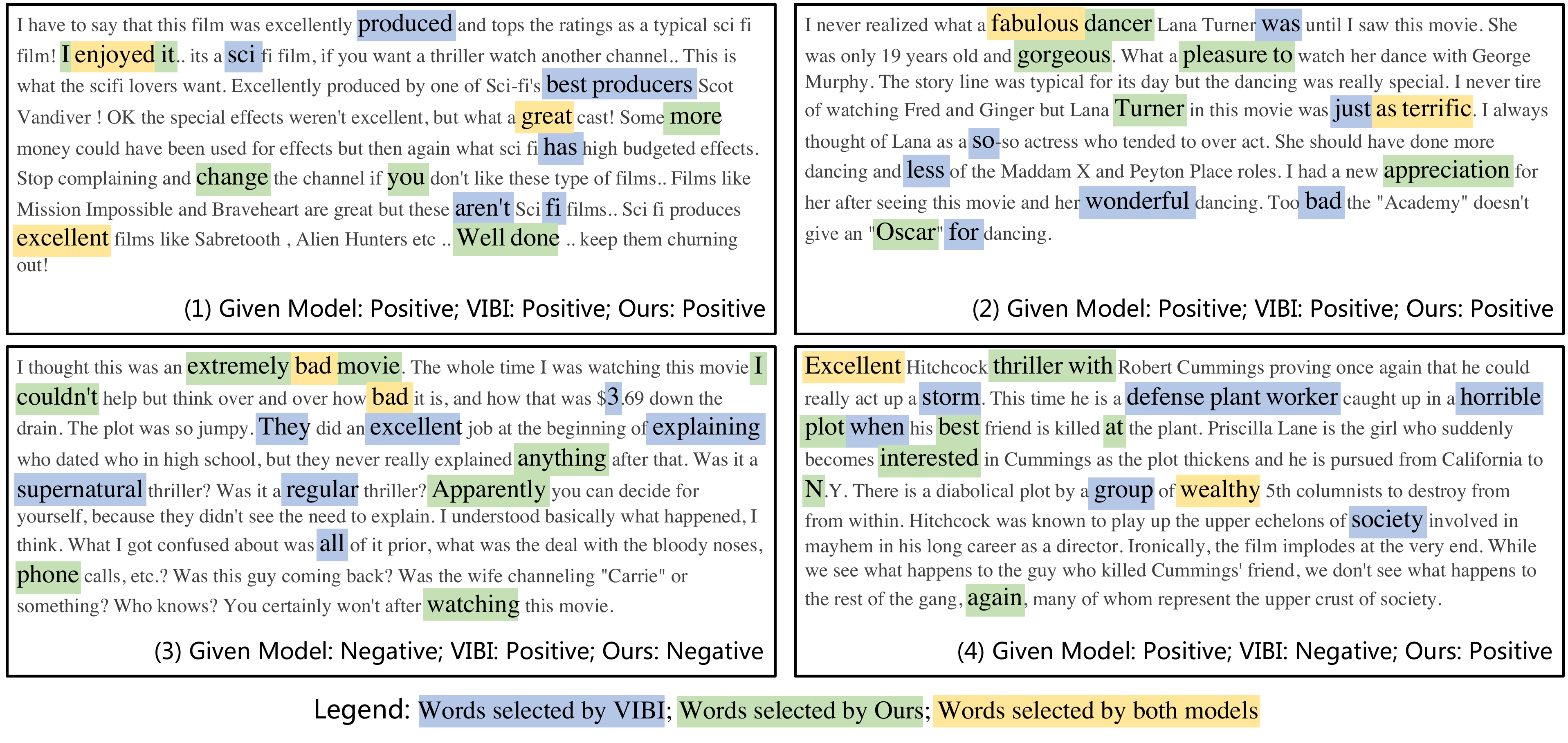}
  \caption{Examples of explanations on the IMDB dataset. The labels predicted by the original given model using all the words, the words selected by the state-of-the-art baseline VIBI, and the words selected by our method are shown, respectively, at the bottom of each panel. Keywords picked by VIBI, our method, and both methods are highlighted in blue, green, and yellow, respectively. In (1) and (2), the given model output consistent predictions using the selected words by both VIBI and ours, whereas (3) and (4), the prediction using full words is inconsistent with that using VIBI's selected words but is still consistent with that using our selected words. Best view in colors.}
  \label{fig:examples_imdb}
\end{figure*}





We conduct comprehensive evaluations on five datasets: 
\begin{itemize}
    \item IMDB sentiment analysis dataset~\cite{maas2011learning}, 
    \item MNIST dataset~\cite{lecun1998gradient} to classify \texttt{3} and \texttt{8}, 
    \item Fashion-MNIST dataset~\cite{xiao2017fashion} to classify \texttt{Pullover} and \texttt{Coat}, 
    \item ImageNet dataset~\cite{deng2009imagenet} to classify \texttt{Gorilla} and \texttt{Zebra}, 
    \item our established mobile sensor dataset from Tencent \emph{Honor of Kings} game for teenager recognition, which we refer to as Tencent Gaming Dataset (TGD) in this paper.
\end{itemize}
The detailed re-organization process for each data will be introduced in the following sections.  

\subsubsection{Methods for Comparison} We compare our method (Ours) with the state-of-the-art model-agnostic baselines: LIME~\cite{ribeiro2016should}, kernel SHAP (SHAP)~\cite{lundberg2017unified}, CXPlain (CXP)~\cite{schwab2019cxplain},
INFD~\cite{yeh2019fidelity},
L2X~\cite{chen2018learning} and VIBI~\cite{bang2019explaining}. We also compare model-specific baselines: Gradient (Grad)~\cite{simonyan2013deep} and Gradient $\times$ Input (GI)~\cite{shrikumar2017learning}.

\subsubsection{Evaluation Metrics}\label{subsec:PE}
{\blue We follow Chen~\cite{chen2018learning} to perform a predictive evaluation for the fidelity of both the selected and the unselected features.} 
For the Fidelity of the Selected features, given an explanation, \emph{e.g.,} selected features $\x_S$, from an arbitrary IFS interpretation method, we evaluate whether the given model $M$ truly use the selected features primarily to generate the very output $\y=M(\x)$. To answer this, we need to approximate $\y$ based on selected features $\x_S$. Thus, we evaluate by the consistency between $\y$ and $M(\tilde{\x}_S)$ (recall that $\tilde{\x}_S$ is $\x$ with unselected features imputed by zeros), denoted by FS-M(\%). {\blue However, $M$ is trained on all features of $\x$, not on $\tilde{\x}_S$~\cite{hooker2019benchmark}. Therefore, we additionally} propose to evaluate the consistency between $\y$ and $A'_s(\tilde{\x}_S)$ as a reference, denoted by FS-A(\%), where $A'_s$ is an trained approximator on $\mathcal{D}$ to learn the mapping $\tilde{\x}_S\rightarrow \y$.
{\blue High FS-M or FS-A result suggests high importance of the selected features.}
Similarly, for the Fidelity of the Unselected features, we evaluate the consistency between $\y$ and $M(\tilde{\x}_{\bar{S}})$, denoted by FU-M(\%); and the consistency between $\y$ and $A'_u(\tilde{\x}_{\bar{S}})$, denoted by FU-A(\%), where $A'_u$ is an trained approximator on $\mathcal{D}$ to learn the mapping $\tilde{\x}_{\bar{S}}\rightarrow \y$. {\blue Low FU-M or FU-A result suggests high importance of the selected features.}
{\blue Note that low FS-A or high FU-A is possible because the number of selected features are usually small. Nonetheless, simultaneous results of high FS-A and low FU-A suggest good selected features.}

For human evaluation, we also follow Chen~\cite{chen2018learning} to evaluate Fidelities of Selected features denoted by FS-H(\%), \emph{i.e.,} whether the predictions made by a human using selected features are consistent with those made by the given model using all the features. {\blue We adopt this metric to evaluate whether human can understand how the given model makes decisions. Note that sometimes human may not understand or be satisfied with features selected by the given model. After all, we are explaining the given model, not human.}

{\blue For the evaluation metric for the fidelities, we report top-1 accuracy (ACC@1), since the five tasks are all binary classification. Specifically, the model outputs are transformed to categorical variables to compute accuracy. We adopt binary masks to select features, \emph{i.e.}, top~$k$ values of $\z=E(\x,\y)$ are set to $1$, others are set to $0$, and then we treat $\x\odot\z$ as the selected features.}
On the other hand, we evaluate the influence of adversarial examples on the feature importance scores by the sensitivity score, SEN(\%), proposed by Yeh~\emph{et al.}~\cite{yeh2019fidelity}. We also report the average explanation Time (by second) Per Sample (TPS) on a single NVidia Tesla m40 GPU. 

\subsubsection{Implementation Details}
In Eq.~\eqref{eq:loss_AIL}, $\ell_s$ adopts cross-entropy. $\ell_u$ adopts cross-entropy for IMDB, MNIST, and TGD, and adopts Wasserstein distance~\cite{lee2019sliced} for Fashion-MNIST and ImageNet. The weights for $\mathcal{L}_u$ in Eq.~\eqref{eq:basic_AIL_relax_pos_final} and~\eqref{eq:basic_AIL_relax_neg_final} are 1e-3 for MNIST and 1 for all other datasets. We adopt the GI method to provide the prior explanations. $\ell_e$ in Eq.~\eqref{eq:constraint} is mean absolute error with the weight to be 1e-3 for ImageNet and 0 for others. $\tau=0.5$ for all the datasets. We constrain the model-capacity of each method to be the same to acquire fair comparisons. For each dataset, we split half test data for validation. For each method on each dataset, we repeat 20 times and report the averaged results. Our implementation uses Keras with Tensorflow~\cite{abadi2016tensorflow} backends. We list all other details in Appendix~\ref{subsec:dt}.




\subsection{IMDB}

The IMDB~\cite{maas2011learning} is a sentiment analysis dataset which consists of 50,000 movie reviews labeled by positive/negative sentiments. Half reviews are for training, and the other half is for testing. We split half of the testing reviews for validation. For the given model, we follow Chen~\emph{et al.}~\cite{chen2018learning} to train a 1D convolutional neural network (CNN) for binary sentiment classification, and achieve the test accuracy of 88.60\%. We develop our approximator with the same architecture as the given model. And we develop our explainer with the 1D CNN used by L2X~\cite{chen2018learning} with a global and a local component. For a fair comparison, each method selects top-$10$ important words as an explanation for a review.



\begin{table}
  \caption{Results on the IMDB dataset.  $^\dagger$ denotes the method uses additional information.}
  \label{tab:imdb}
  \begin{tabular}{lcccccc}
    \toprule
    Method  & FS-M & FU-M & FS-A & FU-A & FS-H & TPS  \\
    \midrule
    Grad   & 85.58  & 87.58 &86.18 & 85.79 & 73.58& 5e-5\\
    GI   & 87.31 & 86.25 & 87.88 & 83.86 & 78.23  &5e-5\\
    LIME   & 89.75 & 82.13 & 88.53 & 82.96  & 83.98 & 3e-2\\
    SHAP   & 50.17 & 99.16 & 50.24 & 99.50 & 53.22  &4e-2\\
    $^\dagger$CXP  & 90.60  & 80.01 & 90.70 & 83.04 & 84.97& 1e-4\\
    INFD  & 40.50 & 99.80 & 64.50& 96.70 & 46.27 & 3e-0\\
    L2X    & 89.23 & 82.90 & 89.05& 83.81&  83.49 & 1e-4\\
    VIBI    &  90.79 & 80.36 & 90.12& 82.57&  84.33& 1e-4\\
    Ours    & {98.48} & {59.05} & {98.70}& {81.83}& {92.98} & 1e-4\\
  \bottomrule
\end{tabular}
\end{table}

As shown in Table~\ref{tab:imdb}, our method significantly outperforms state-of-the-art baseline methods. Especially, our FS-M score shows nearly optimal fidelity, which is objectively validated by the original given model. Given that our FU-A score is similar to those of baselines, which shows that our selected features are indeed important, we demonstrate that the effectiveness and superiority of our method are significant. We present some examples of selected words of our method and the state-of-the-art baseline VIBI in Fig.~\ref{fig:examples_imdb}. As shown in Fig.~\ref{fig:examples_imdb}, our method not only selects more accurate keywords, but also provides more interpretable word combinations, such as ``I enjoyed it'', ``fabulous dancer'', ``extremely bad movie'', and ``excellent thriller''. Even though ``I'', ``it'', ``dancer'', ``movie'', and ``thriller'' are not apparent whether they are positive or negative words. Especially in Fig.~\ref{fig:examples_imdb} (2), our method picks the word ``Oscar'', which is not explicit positive, but its underlying logic suggests positive sentiment. These inspiring examples support the significant superiority of our method. 

\subsubsection{Human Evaluation}\label{subsubsubsec:he_imdb}
We also evaluate with the help of humans to quantify how interpretable are those selected words. We randomly select $500$ reviews in the testing set for this human evaluation. We invite $20$ Tencent employees to infer the sentiment of a review given only the selected words. The explanations provided by different interpretation methods are randomly mixed before sent to these employees. The final prediction of each review is averaged over multiple human annotations. For the explanations that are difficult to infer the sentiments, we ask the employees to provide random guesses. Finally, as shown by the FS-H scores in Table~\ref{tab:imdb}, our method significantly outperforms baseline methods as well.

\subsubsection{Ablation Study}\label{subsubsubsec:ab_imdb}

\begin{figure}[h]
  \centering
  \includegraphics[width=0.8\linewidth]{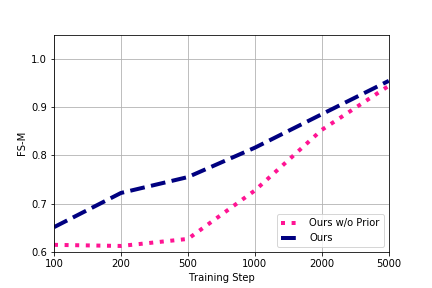}
  \caption{The FS-M scores of our method with and without the prior knowledge-based warm start.  }
  \label{fig:warm_start}
\end{figure}

We evaluate three variants of our method by ablating our three components, \emph{i.e.,}
the model output feedback (Output), AIL, and prior knowledge-based warm start (Prior). In Table~\ref{tab:imdb_ab}, we show the effectiveness of both the model output feedback and AIL. It is worthy of mentioning that, although the warm start strategy does not improve the final scores, it boosts the convergence rate at the start of optimization, as shown in Fig.~\ref{fig:warm_start}.

\subsubsection{Sanity Check}\label{subsubsubsec:sc_imdb}

We perform the model and data randomization tests suggested by Adebayo~\emph{et al.}~\cite{adebayo2018sanity}. We evaluate the sanity by the cosine correlation between binary masks, \emph{i.e.,} the original mask, and the other one resulted from randomization. The sanity scores for model and data randomization tests are 9.39\% and 10.25\%, respectively, which shows that our explanations are dependent on models and then are valid.

\begin{table}
  \caption{Results of the ablation study.}
  \label{tab:imdb_ab}
  \begin{tabular}{lcccc}
    \toprule
    Method  & FS-M & FU-M & FS-A & FU-A  \\
    \midrule
    Ours    & 98.48 & 59.05 & 98.70& 81.83\\
     w/o Output   & 92.47 & 54.28 & 91.99& 79.82\\
    w/o AIL     & 78.31 & 97.62 & 99.33& 94.42\\
    w/o Prior    & 98.38 & 60.22 & 98.97& 81.07\\
  \bottomrule
\end{tabular}
\end{table}

\subsection{MNIST}

\begin{figure}[h]
  \centering
  \includegraphics[width=\linewidth]{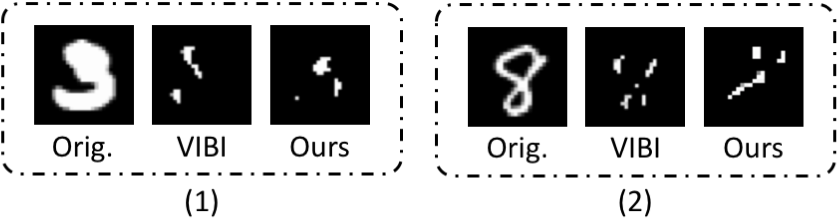}
  \caption{Examples of explanations on the MNIST dataset. In each panel, an original image and the masked images by VIBI and our method are presented from left to right. The labels inferred by the original model using the full images are \texttt{3} and \texttt{8} for panels (1) and (2), respectively.}
  \label{fig:examples_mnist}
\end{figure}

We select the data of \texttt{3} and \texttt{8} of the MNIST dataset~\cite{lecun1998gradient} for binary classification with 11, 982 training and 1984 testing images. 
We train a 2D CNN with two convolutional layers for the classification and achieve the test accuracy of 99.89\%. We develop our approximators are by the same architecture as the given model, whereas we develop our explainer by a 2D CNN with three convolutional layers only. Each method selects top-$25$ pixels as an explanation for an image.

\begin{table}
  \caption{Results on the MNIST dataset.}
  \label{tab:mnist}
  \begin{tabular}{lccccccc}
    \toprule
    Method  & FS-M & FU-M & FS-A & FU-A & SEN & FS-H & TPS  \\
    \midrule
    Grad   & 98.19 & 68.75  & 99.55 & 99.55 & 139 & 94.37 & 5e-5\\
    GI   & 99.45 & 67.64  &99.55 &99.24  & 120  & 94.58  & 5e-5\\
    LIME   & 80.37 & 99.75 & 82.46& 99.80 &62.6  & 70.29 & 3e-2\\
    SHAP   & 92.74 & 90.83  & 98.87 & 99.75 & 87.2 & 87.75 & 4e-2 \\
    $^\dagger$CXP   & 99.40 & 64.77 & 99.70 & 99.24 & 91.8& 94.57& 1e-4\\
    INFD  & 89.62 & 96.62 & 99.95 & 99.95 & 101 & 84.33& 1e-0\\
    L2X   & 91.38 & 91.18 & 98.54  & 99.65 & 6.90 & 86.58&1e-4\\
    VIBI    & 98.29  & 86.29 & 99.29  & 99.65 & 6.35  & 92.17 &1e-4\\
    Ours    & 99.04 & 74.70 & 99.80 &  99.70 &  6.11 & 94.46 &1e-4\\
  \bottomrule
\end{tabular}
\end{table}

As shown in Table~\ref{tab:mnist}, our method still outperforms state-of-the-art model-agnostic baseline methods except for the CXPlain method, which uses the additional true label for each sample and is highly sensitive to adversarial examples. The Grad and GI methods are model-specific and not robust when facing challenging data, \emph{e.g.,} see Tables~\ref{tab:imdb},~\ref{tab:fmnist}, and~\ref{tab:imagenet}.
Compared with the next-best model-agnostic baseline VIBI, the strength of our method determined by the FU-M score is to select the necessary features of a sample for recognition. 
We show some examples of selected pixels of our method and VIBI in Fig.~\ref{fig:examples_mnist}. As shown in Fig.~\ref{fig:examples_mnist} (1), the VIBI masked image is closer to \texttt{8} than \texttt{3}, whereas our masked image is more similar to \texttt{3}. In Fig.~\ref{fig:examples_mnist} (2), though the VIBI masked image is similar to \texttt{8}, but it is also close to a \texttt{3}. In contrast, our masked image can never be \texttt{3}. Since we are interpreting the recognition logic of model rather than that of humans, it is important to select features in favor of the machine logic, \emph{e.g.,} considering both possibility and impossibility.

\subsubsection{Human Evaluation}\label{subsubsubsec:he_mnist}
We randomly select $500$ images in the testing set for this human evaluation. We invite $15$ Tencent employees who are experts in machine learning to perform the same binary classification given only the masked images. Specifically, we ask the subjects to provide two scores in the range of $[0,1]$ for each image. Each score is for the possibility of each class (\texttt{3} or \texttt{8}). We perform $\ell_1$ normalization for the scores as the final prediction probabilities. Other settings and procedures are similar to Section~\ref{subsubsubsec:he_imdb}. Finally, as shown by the FS-H scores in Table~\ref{tab:mnist}, our method significantly outperforms baseline methods as well.

\subsection{Fashion-MNIST}

\begin{figure}[t]
  \centering
  \includegraphics[width=0.8\linewidth]{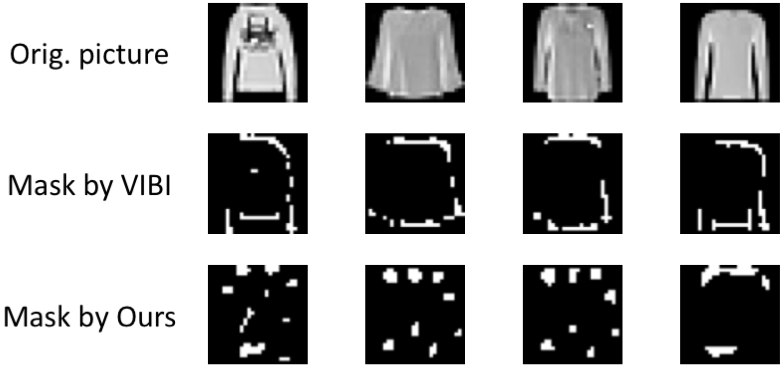}
  \caption{Examples of explanations on the Fashion-MNIST dataset. The first, second and third lines list the original images, the images masked by VIBI, the images masked by our method, respectively.}
  \label{fig:examples_fmnist}
\end{figure}

\begin{table}
  \caption{Results on the Fashion-MNIST dataset.}
  \label{tab:fmnist}
  \begin{tabular}{lcccccc}
    \toprule
    Method  & FS-M & FU-M & FS-A & FU-A & SEN & TPS  \\
    \midrule
    Grad   & 58.60 & 76.20 & 93.70 &95.45  & 2528  & 5e-5 \\
    GI   & 62.15 & 66.25 & 93.35 & 94.45 & 2662 &  5e-5\\
    LIME   & 75.63 & 94.30 & 73.60& 97.35 & 61.03& 3e-2 \\
    SHAP & 63.29  & 55.78  & 93.97 & 95.58 & 84.59    &  4e-2\\
    $^\dagger$CXP   & 59.65 & 16.50 & 94.85 & 95.10 & 107 & 1e-4\\
    INFD  & 100.0 & 45.80 & 100.0 & 100.0 & 87.37 & 5e-1 \\
    L2X   & 77.30 & 87.30 & 89.85  & 96.00 & 1.76 &  1e-4\\
    VIBI    & 84.10  & 70.85 & 91.90  & 94.40 & 17.36 & 1e-4 \\
    Ours    & 97.80 & 66.65 & 99.80 &  99.65 &  0.70 &  1e-4 \\
  \bottomrule
\end{tabular}
\end{table}

\begin{figure}[t]
  \centering
  \includegraphics[width=\linewidth]{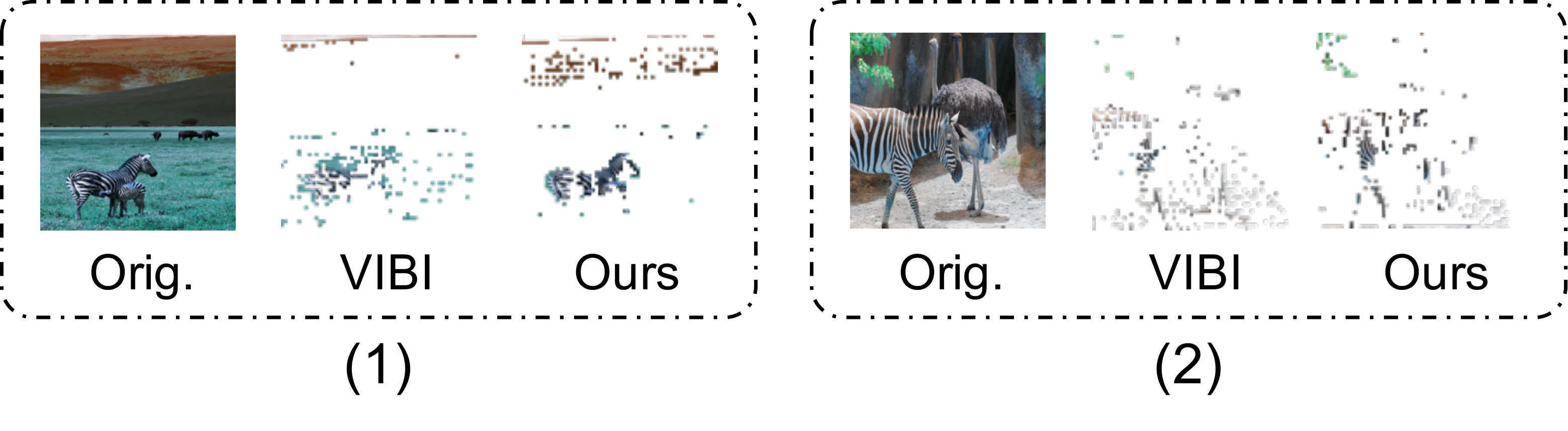}
  \caption{Examples of explanations on the ImageNet dataset. In each panel, an original image and the masked images by VIBI and our method are presented from left to right. Best view in colors.}
  \label{fig:examples_imagenet}
\end{figure}

The Fashion-MNIST dataset~\cite{xiao2017fashion} is a dataset of Zalando's article images, which consists of $28\times 28$ images of $10$ classes. We select the data of \texttt{Pullover} and \texttt{Shirt} for binary classification dataset with 12,000 training and 2000 testing images. 
We train a 2D CNN with the same architecture as that for MNIST for the classification and achieve the test accuracy of 92.20\%. The architectures of our approximators and explainer are also the same as those for MNIST. Each method selects top-$64$ pixels as an explanation for an image.

As shown in Table~\ref{tab:fmnist}, our method outperforms state-of-the-art baseline methods except for the INFD method. Since INFD adds perturbation to each feature and directly performs regression, it is suitable for well-aligned data, \emph{e.g.,} the Fashion-MNIST dataset. However, as shown in Tables~\ref{tab:imdb} and~\ref{tab:mnist}, INFD's performances are disappointing for data that are not well-aligned. Therefore, our method is more robust. Moreover, INFD is extremely time-consuming to be applied to practical applications and is sensitive to adversarial examples. Compared with the next-best baseline VIBI, we show some examples of selected pixels in Fig.~\ref{fig:examples_fmnist}. As shown in Fig.~\ref{fig:examples_fmnist}, VIBI primarily focuses on the contours, whereas our method focuses on relatively fixed local regions. Since the data are well-aligned, the explanations provided by our method are more consistent with the machine logic of the original model.

\subsection{ImageNet}

\begin{figure}[t]
  \centering
  \includegraphics[width=\linewidth]{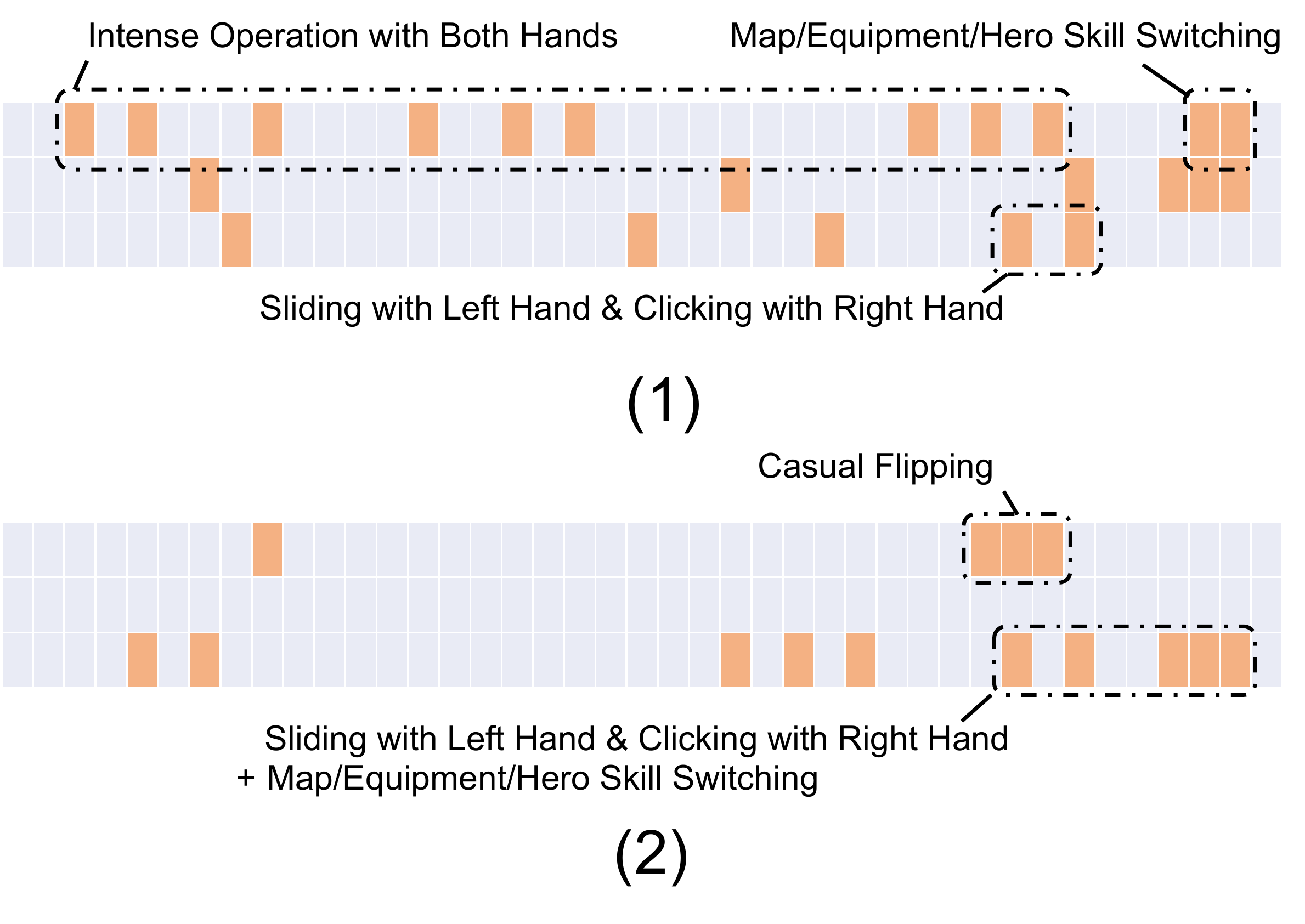}
  \caption{Examples of explanations on the TGD dataset (Honor of Kings). Each panel shows the mask for each time-series data sample with the time dimension of $3$. (1) and (2) show samples of a teenager and an adult, respectively. Selected features are in orange. Best view in colors.}
  \label{fig:examples_thk}
\end{figure}

\begin{table}
  \caption{Results on the ImageNet dataset.}
  \label{tab:imagenet}
  \begin{tabular}{lcccccc}
    \toprule
    Method  & FS-M & FU-M & FS-A & FU-A & SEN & TPS  \\
    \midrule
    Grad   & 55 & 91 & 66& 99 & 1e+6 & 2e-3 \\
    GI   & 56 & 95 &74 &89  & 1e+6 & 2e-3 \\
    LIME   & 77  & 98 & 72 & 96 & 76.1 & 1e-0  \\
    SHAP   & 66 & 96 & 75 & 90 & 89.4 & 2e-0 \\
    $^\dagger$CXP  & 77 & 96 &61 &95  & 41.38  & 5e-3 \\
    INFD  & - & - &- &-  & - & 4e+3\\
    L2X   & 78 & 99 & 75  & 96 & 41.32 &  5e-3\\
    VIBI    & 78  & 98 & 78  & 96 & 40.10 & 5e-3 \\
    Ours    & 83 & 98 & 90 &  93 &  39.91 &  5e-3\\
  \bottomrule
\end{tabular}
\end{table}

We select the data of \texttt{Gorilla} and \texttt{Zebra} from ImageNet~\cite{deng2009imagenet} for binary classification.
We adopt the MobileNet~\cite{howard2017mobilenets} and train only the top layer for the classification and achieve the test accuracy of 100\%. We develop our approximators with the same architecture and adopt the U-Net~\cite{ronneberger2015u} for our explainer.
Each method selects top-10\% pixels as an explanation for an image. As shown in Table~\ref{tab:imagenet}, our method outperforms state-of-the-art baselines.  We exhibit some examples of selected pixels in Fig.~\ref{fig:examples_imagenet} and compare them with the best baseline VIBI. As shown in Fig.~\ref{fig:examples_imagenet} (1), our selected pixels are more concentrated on the label-related regions, which demonstrates that our method can improve the model identifiability. Fig.~\ref{fig:examples_imagenet} (2) shows that our method can better avoid irrelevant regions, \emph{e.g.,} the ground and the back of an ostrich.

\subsection{TGD}

Last, we apply our method to the Tencent Gaming Dataset (TGD), which consists of 100 million samples from 5 million gamers. Each sample is a $3\times 643$ time-series data with the time dimension and feature dimension of $3$ and $643$, respectively. We extract the features from inertia sensors and touch information of mobile phones, in both time and frequency domains, and categorize in $41$ groups. Each feature vector of a sample corresponds to a $2$-second operation during the game. Three vectors are ordered by time, but not necessarily continuous in time. The learning task is the teenage gamer (age $\leq17$) recognition. 
The original model is a stacked LSTM with an accuracy of 90.16\%. The approximator uses the same structure, and the explainer is also a stacked LSTM. Our method achieves the FS-M, FU-M, FS-A, FU-A, and SEN scores of 95.68\%, 82.24\%, 95.33\%, 82.37\%, and 0.18\%, respectively, selecting only 10\% of features. We show examples of selected features in Fig.~\ref{fig:examples_thk}. In Fig.~\ref{fig:examples_thk} (1), the teenage gamer performs a complex operation excitedly at the start but performs a monotonous/regular operation at the end. Whereas, in Fig.~\ref{fig:examples_thk} (2), the adult gamer starts with casual flipping of the mobile phone, and ends with a complex/skilled operation.


\section{Conclusion}\label{sec:conclusion}

In this paper, we investigate the model interpretation problem in the favored direction of Instance-wise Feature Selection (IFS). We propose a Model-agnostic Effective Efficient Direct (MEED) IFS framework for model interpretation. Specifically, we consider the model output feedback as an additional input to learn an explainer to mitigate the sanity and information transmission problems. Furthermore, we propose an adversarial infidelity learning (AIL) mechanism to screen relative unimportant features for mitigating the combinatorial shortcuts and the model identifiability problems. Our theoretical analyses show that AIL can mitigate the model identifiability and combinatorial shortcuts problems. Our experimental results reveal that AIL can mitigate the model identifiability problem and learn more necessary features. Moreover, {\blue our extension to integrate efficient interpretation methods as proper priors has been shown to provide a warm start and mitigate the information transmission problem.} Comprehensive empirical evaluation results provided by quantitative metrics and human evaluation demonstrate the effectiveness, superiority, and robustness of our method.

\begin{acks}
Jian Liang, Bing Bai, Yuren Cao, and Kun Bai would like to acknowledge the support from the TuringShield team of Tencent. Fei Wang would like to acknowledge the support from AWS Machine Learning for Research Award and Google Faculty Research Award.
\end{acks}

\bibliographystyle{ACM-Reference-Format}
\bibliography{main}

\newpage
\appendix

\section{Proofs and Derivations}

\subsection{Proof of Theorem~\ref{th:mi}}\label{subsec:proof_1}
\begin{proof}
The proof follows that of Theorem 1 of Chen~\emph{et al.}~\cite{chen2018learning}. 

(1) Forward direction: Given the definition of $S^*$, we have for any pair $(\x,\y)$, and any explainer $E: S\mid \x,\y$,
\begin{equation}
\begin{split}
\mathbb{E}[\log p(\y\mid \x_S)- \log p(\y\mid \x_{\bar{S}})] 
\leq\mathbb{E}[\log p(\y\mid \x_{S^*})- \log p(\y\mid \x_{\bar{S}^*})]
\end{split}
\end{equation}
In the case when $S^*$ is a set instead of a singleton, we identify $S^*$ with any distribution that assigns arbitrary probability to each elements in $S^*$, and with zero probability outside $S^*$. With abuse of notation, $S^*(\x,\y)$ indicates both the set function that maps every pair $(\x,\y)$ to a set $S^*$ and any real-valued function that maps $(\x,\y)$ to an element in $S^*$.
Taking expectation over the distribution of $(\x,\y)$, and adding $\mathbb{E}[\log p(\y)]$ at both sides, we have
\begin{equation}
I(\x_S;\y) - I(\x_{\bar{S}};\y) \leq I(\x_{S^*};\y)-I(\x_{\bar{S}^*};\y)
\end{equation}
for any explainer $E: S\mid \x,\y$.

(2) Reverse direction: The reverse direction is proved by contradiction. 
Since the optimal explanation $S \mid \x,\y$ satisfies
\begin{equation}\label{eq:proof_1_0}
I(\x_{S'};\y) - I(\x_{\bar{S}'};\y) \leq I(\x_{S};\y)-I(\x_{\bar{S}};\y)
\end{equation}
for any other $S'\mid \x,\y$, assume the optimal explanation $S \mid \x,\y$ is such that there exists a set $\mathcal{S}$ of
nonzero probability, over which $S\mid \x,\y$ does not degenerates to an element in $S^*$. Concretely, we define $\mathcal{S}$ as
\begin{equation}
\mathcal{S} = \{\x,\y:p(S \notin S^*\mid \x,\y)>0\}.
\end{equation}
For any $(\x,\y)\in\mathcal{S}$, we have
\begin{equation}\label{eq:proof_1_1}
\begin{split}
\mathbb{E}[\log p(\y\mid \x_S)- \log p(\y\mid \x_{\bar{S}})] 
<\mathbb{E}[\log p(\y\mid \x_{S^*})- \log p(\y\mid \x_{\bar{S}^*})],
\end{split}
\end{equation}
where $S^*(\x,\y)$ is a deterministic function in the set of distributions that assign arbitrary probability to each elements in $S^*$, and with zero probability outside $S^*$. Outside $\mathcal{S}$, we always have
\begin{equation}\label{eq:proof_1_2}
\begin{split}
\mathbb{E}[\log p(\y\mid \x_S)- \log p(\y\mid \x_{\bar{S}})] 
\leq\mathbb{E}[\log p(\y\mid \x_{S^*})- \log p(\y\mid \x_{\bar{S}^*})]
\end{split}
\end{equation}
from the definition of $S^*$. As $\mathcal{S}$ is of nonzero size over $p(\x,\y)$, combining Eq.~\eqref{eq:proof_1_1} and Eq.~\eqref{eq:proof_1_2}, taking expectation with respect to $p(\x,\y)$ and adding $\mathbb{E}[\log p(\y)]$ at both sides, we have
\begin{equation}
I(\x_S;\y) - I(\x_{\bar{S}};\y) < I(\x_{S^*};\y)-I(\x_{\bar{S}^*};\y)
\end{equation}
which is a contradiction to Eq.~\eqref{eq:proof_1_0}.
\end{proof}

\subsection{Derivations for The Variational Lower Bounds}\label{subsec:VLB}

\begin{proof}
First, for selected features, we have:
\begin{equation}
\begin{split}
    I(\x_S;\y) &= \mathbb{E}\biggl[\log\frac{p(\x_S,\y)}{p(\x_S)p(\y)}\biggr]
    = \mathbb{E}\biggl[\log\frac{p(\y\mid\x_S)}{p(\y)}\biggr]\\
    &= \mathbb{E}[\log p(\y\mid\x_S)] + \mbox{Const.} \\
    &=\mathbb{E}_{\x}\mathbb{E}_{\y\mid\x}\mathbb{E}_{S\mid\x,\y}\mathbb{E}_{\y\mid\x_S}[\log p(\y\mid\x_S)] + \mbox{Const.}
\end{split}
\end{equation}
For any $q_s(\y\mid \x_S)$, we obtain the lower bound by applying the Jensen's inequality:
\begin{equation}
\begin{split}
\mathbb{E}_{\y\mid\x_S}[\log p(\y\mid\x_S)]
&\geq \int [\log q_s(\y\mid \x_S)]dp(\y\mid\x_S)\\
&= \mathbb{E}_{\y\mid\x_S}[\log q_s(\y\mid \x_S)]. 
\end{split}
\end{equation}

It is similar for unselected features.
\end{proof}

\subsection{Proof of Theorem~\ref{th:ail}}\label{subsec:proof_2}

\begin{proof}
For the problem in Eq.~\eqref{eq:basic_AIL_relax_pos_final} and~\eqref{eq:basic_AIL_relax_neg_final},
$A_u$ learns $p(\y\mid \x_{\bar{S}},\mathbf{1}-\v)$, where $\mathbf{1}$ is a vector with all the elements being $1$. 

Therefore, our AIL mechanism learns $\v$ to minimize $p(\y\mid \x_{\bar{S}},\mathbf{1}-\v)$. Since
\begin{equation}\label{eq:mi_soft}
\begin{split}
    &I((\x_{\bar{S}},\mathbf{1}-\v);\y) \\
    &= \mathbb{E}\biggl[\log\frac{p(\x_{\bar{S}},\mathbf{1}-\v,\y)}{p(\x_{\bar{S}},\mathbf{1}-\v)p(\y)}\biggr]\\
    &= \mathbb{E}\biggl[\log\frac{p(\y\mid\x_{\bar{S}},\mathbf{1}-\v)}{p(\y)}\biggr]\\
    &= \mathbb{E}[\log p(\y\mid\x_{\bar{S}},\mathbf{1}-\v)] + \mbox{Const.} \\
    &=\mathbb{E}_{\x}\mathbb{E}_{\y\mid\x}\mathbb{E}_{\v\mid\x,\y}\mathbb{E}_{\y\mid\x_{\bar{S}},\mathbf{1}-\v}[\log p(\y\mid\x_{\bar{S}},\mathbf{1}-\v)] \\
    & \ \ \ \ + \mbox{Const.},
\end{split}
\end{equation}
the mutual information $I((\x_{\bar{S}},\mathbf{1}-\v);\y)$ is minimized. By the property of mutual information, we assume that the minimization of $I((\x_{\bar{S}},\mathbf{1}-\v);\y)$ encourages the independence between $(\x_{\bar{S}},\mathbf{1}-\v)$ and $\y$, which leads to  
\begin{equation}\label{eq:mi_soft_ind}
\begin{split}
p(\x_{\bar{S}},\mathbf{1}-\v,\y)
=p(\x_{\bar{S}},\mathbf{1}-\v)p(\y)
\end{split}
\end{equation}
Thus, by marginalizing $\x_{\bar{S}}$ at both sides, we have
\begin{equation}\label{eq:mi_soft_ind_marg}
\begin{split}
\int_{\x_{\bar{S}}}p(\x_{\bar{S}},\mathbf{1}-\v,\y)
&=\int_{\x_{\bar{S}}}p(\x_{\bar{S}},\mathbf{1}-\v)p(\y)\\
&\Rightarrow
p(\mathbf{1}-\v,\y)
=p(\mathbf{1}-\v)p(\y).
\end{split}
\end{equation}
Therefore, the independence between $\mathbf{1}-\v$ and $\y$ is encouraged as well. Since $\mathbf{1}-\v$ and $\v$ are deterministic between each other, then for any set $\mathcal{S}_1$ such that $\v\in \mathcal{S}_1$, there exists a fixed set $\mathcal{S}_2$ such that $\mathbf{1}-\v\in \mathcal{S}_2$, and vice versa. Thus we have for any set $\mathcal{S}_1$, 
\begin{equation}\label{eq:mi_soft_ind_deter}
\begin{split}
p(\v\in\mathcal{S}_1,\y) 
&= p(\mathbf{1}-\v\in\mathcal{S}_2,\y) \\
&=p(\mathbf{1}-\v\in\mathcal{S}_2)p(\y)
= p(\v\in\mathcal{S}_1)p(\y).
\end{split}
\end{equation}
Thus, the independence between $\v$ and $\y$ is also encouraged. 
\end{proof}

\subsection{Derivation for Eq.~\eqref{eq:naive_bayes}}\label{subsec:proof_3}

\begin{proof}
\begin{equation} 
\begin{split}
&p(\delta = j\mid \x,M,E,H) \\
&= \frac{p(\delta = j, E,H\mid \x,M)}{p( E,H\mid \x,M)}\\
&=\frac{p(\delta = j\mid \x,M) p( E,H\mid\delta = j, \x,M)}{p( E,H\mid \x,M)}\\
&=\frac{p(\delta = j\mid \x,M) p( E\mid\delta = j, \x,M)p( H\mid\delta = j, \x,M)}{p( E,H\mid \x,M)}\\
&=\frac{p(\delta = j\mid \x,M) \frac{p( E,\delta = j\mid \x,M)}{p(\delta = j\mid \x,M)}\frac{p( H,\delta = j\mid \x,M)}{p(\delta = j\mid \x,M)}}{p( E,H\mid \x,M)}\\ 
&=\frac{p(\delta = j\mid \x,M) \frac{p( E\mid \x,M)p( \delta = j\mid \x,M,E)}{p(\delta = j\mid \x,M)}\frac{p( H\mid\x,M) p( \delta = j\mid \x,M,H)}{p(\delta = j\mid \x,M)}}{p( E,H\mid \x,M)}\\ 
&=C(E,H\mid\x,M)\frac{p( \delta = j\mid \x,M,E)p( \delta = j\mid \x,M,H)}{p(\delta = j\mid \x,M)},
\end{split}
\end{equation}
where
\begin{equation}
 C(E,H\mid\x,M) = \frac{p( E\mid \x,M)p( H\mid \x,M)}{p( E,H\mid \x,M)}. 
\end{equation}
The third equation is from the assumption of conditional independence between the interpretation models $E$ and $H$ given $\delta,\x$ and $M$.
Assuming $p(\delta = j\mid \x,M) = \frac{1}{d}$ for all $j\in[d]$, because we have no knowledge of the explainer, then we have
\begin{equation} 
\begin{split}
p(\delta = j\mid \x,M,E,H) 
&= \frac{p(\delta = j\mid \x,M,E,H) }{\sum_{j'=1}^d
p(\delta = j'\mid \x,M,E,H)}\\
&=\frac{p(\delta = j\mid \x,M,E)p(\delta = j\mid \x,M,H)}{\sum_{j'=1}^dp(\delta = j'\mid \x,M,E)p(\delta = j'\mid \x,M,H)}. 
\end{split}
\end{equation}
\end{proof}

\section{Implementation Details}\label{subsec:dt}

\subsection{Details for Our Method}\label{subsec:dt_our}

For $\ell_u$ with cross entropy loss, in Eq.~\eqref{eq:basic_AIL_relax_neg_final} we still minimize $\mathcal{L}_u$, and just replace the target $\y$ for $\ell_u$ by $\mathbf{1}-\y$, following the suggestion of the Relativistic GAN~\emph{et al.}~\cite{jolicoeur2018relativistic}.
For $\ell_u$ with the Sliced Wasserstein distance~\cite{lee2019sliced}, the number of random vectors is $128$ for Fashion-MNIST and $256$ for ImageNet.

For optimizers, we use RMSprop for IMDB and Adadelta for image datasets, with the default hyperparameters. The learning rates are fixed.
For TGD, we use Adam with learning rate of $2e-4$, $\beta_1=0.5,\beta_2=0.999$, with the learning-rate decay of 30\% for every 50,000 steps. The batch size is $32$ for IMDB and ImageNet, $128$ for MNIST and Fashion-MNIST, and $1024$ for TGD. The hyper-parameter tuning set for both $\mathcal{L}_u$ and $\mathcal{L}_e$ is $\{0,10^{-3},10^{-2},10^{-1},1\}$.

For IMDB, we adopt the structure of the original model of L2X~\cite{chen2018learning} for the same structure for our original model and approximators. We adopt adopt the structure of the explainer of L2X~\cite{chen2018learning} for the structure for our explainer, with a global and a local component.
For MNIST and Fashion-MNIST, the structure of our original model and approximators is shown in Table~\ref{tab:simple_CNN}. Whereas the structure for our explainer is shown in Table~\ref{tab:simple_CNN_exp}.
For ImageNet, we adopt the MobileNet module in the package of keras.applications.mobilenet without the top layer as our explainer. The model parameters pretrained on ImageNet are fixed. We only stack a global max-pooling layer and learn a full-connected top layer. We adopt the preprocess\_input function in the keras.applications.mobilenet package for image pre-processing. We perform a max-pooling of kernel size and stride of $4$, before generate the feature important scores, and perform an up-sampling with kernel size and stride of $4$ when masking pixels. We adopt the adopt the U-Net~\cite{ronneberger2015u} for our explainer, whose structure is complex and then omitted due to space limitations. Similarly, the structures of our modules for TGD are also omitted. Readers may refer to the publicly-available code for more implementation details.

For IMDB, the model output $\y$ is input to a MLP with three hidden layers with $100$ neurons and the ReLu activation, before being concatenated to the global component of the explainer. For image datasets, the model output $\y$ is linearly mapped to the same shape with the first channel of an image and is concatenated to the raw image as an additional channel. For TGD, the model output $\y$ is linearly mapped to the same shape as an raw data sample and concatenated to the data sample in the feature dimension.

\setlength{\tabcolsep}{2pt}
\begin{table}[htpb]\small
\caption{The CNN structure for our original model and approximators for MNIST and Fashion-MNIST.}\label{tab:simple_CNN}
\begin{center}
\begin{tabular}{c|c|c|c|c|c}
\hline\hline
 Layer & \# Filters &  Kernel Size & Stride & \# Padding &Activation \\\hline\hline
 Convolution & 32 &  3 & 3 & default & ReLu\\\hline
 Convolution & 64 &  3 & 3 & default & ReLu\\\hline
 Max-pooling & - &  2 & 2 & default & -\\\hline
 Dropout(0.25) & - &  - & - & - & -\\\hline
 Flatten & - &  - & - & - & -\\\hline
 Fully-Connected & 128 &  - & - & - & ReLu\\\hline
 Dropout(0.5) & - &  - & - & - & -\\\hline
 Fully-Connected & 2 &  - & - & - & Softmax\\\hline
 \hline
\end{tabular}
\end{center}
\end{table}

\setlength{\tabcolsep}{2pt}
\begin{table}[htpb]\small
\caption{The CNN structure for our explainer for MNIST and Fashion-MNIST.}\label{tab:simple_CNN_exp}
\begin{center}
\begin{tabular}{c|c|c|c|c|c}
\hline\hline
 Layer & \# Filters &  Kernel Size & Stride & \# Padding &Activation \\\hline\hline
 Convolution & 32 &  3 & 3 & same & ReLu\\\hline
 Convolution & 64 &  3 & 3 & same & ReLu\\\hline
 Convolution & 1 &  3 & 3 & same & Linear\\\hline
 \hline
\end{tabular}
\end{center}
\end{table}

\subsection{Details for Baseline Methods}\label{subsec:dt_other}

For Grad, we compute the gradient of the selected class with respect to the input feature and uses the absolute values as importance scores. We perform summation operations to form the importance scores with proper shapes. For GI, the gradient is multiplied by the input feature before calculate the absolute value. For INFD, we select the Noisy Baseline method for consistent comparisons, since its another method Square is only suitable for image datasets. The structures of explainers are the same for CXP, L2X, VIBI, and Ours. The structures of original models and approximators are the same for L2X, VIBI, and Ours. 

The hyper-parameters of each method are tuned according the strategy mentioned in their respective papers.

On ImageNet, for all the baseline methods, we perform a max-pooling of kernel size and stride of $4$ for the feature important scores, and perform an up-sampling with kernel size and stride of $4$ when masking pixels.

\end{document}